\documentclass[lettersize,journal]{IEEEtran}
\usepackage{amsmath,amsfonts}
\usepackage{algorithmic}
\usepackage{algorithm}
\usepackage{array}
\usepackage[caption=false,font=normalsize,labelfont=sf,textfont=sf]{subfig}
\usepackage{textcomp}
\usepackage{stfloats}
\usepackage{url}
\usepackage{verbatim}
\usepackage{graphicx}
\usepackage{multirow}
\usepackage{cite}
\usepackage{xcolor}
\usepackage{booktabs}
\definecolor{qmcolor}{RGB}{0,0,0}
\newcommand{\zqm}[1]{{\color{qmcolor}#1}}
\definecolor{zccolor}{RGB}{0,0,0}
\newcommand{\zc}[1]{{\color{zccolor}#1}}
\definecolor{revred}{RGB}{0,0,0}
\newcommand{\rev}[1]{{\color{revred}#1}}

\begin{document}

\title{Na-IRSTD: Enhancing Infrared Small Target Detection via Native-Resolution Feature Selection and Fusion}

\author{Qian Xu, Chi Zhang, Qiming Zhang, Xi Li, Haojuan Yuan, Mingjin Zhang, ~\IEEEmembership{Member, IEEE}}



\maketitle

\begin{abstract}
Infrared small target detection (IRSTD) faces the inherent challenge of precisely localizing dim targets amid complex background clutter. \zqm{While progress has been made, existing methods usually follow conventional strategies to downsample features and discard small targets' details, resulting in suboptimal performance.} In this paper, we present Na-IRSTD, a native-resolution feature extraction and fusion framework for IRSTD. This framework elegantly incorporates native-resolution features to preserve subtle target cues, overcoming the resolution limitations of existing infrared approaches and significantly improving the model's ability to localize small targets. \zqm{We also introduce an effective token reduction and selection strategy, which selects target patches with high accuracy and confidence, boosting the low-level details of the feature while effectively reducing  native-resolution patch tokens compared to dense processing, thereby avoiding imposing an unbearable computational burden.} Extensive experiments demonstrate the robustness and effectiveness of our token reduction and selection strategy across multiple public datasets. Ultimately, our Na-IRSTD model achieves state-of-the-art performance on four benchmarks. 
\end{abstract}


\section{Introduction}
\zc{The objective of infrared small target detection (IRSTD) is to precisely identify and locate diminutive targets in infrared imagery. Such targets are often characterized by their minimal pixel occupancy and inherent low contrast when set against complex backgrounds. This \zqm{task} proves indispensable in vital, safety-critical domains, including maritime surveillance, remote sensing, and environmental monitoring~\cite{kou2023infrared,rawat2020review,8307251,zhu2025shifting,ma2025temporal}}.
\begin{figure}[t!]
    \centering
    \includegraphics[width=1\linewidth]{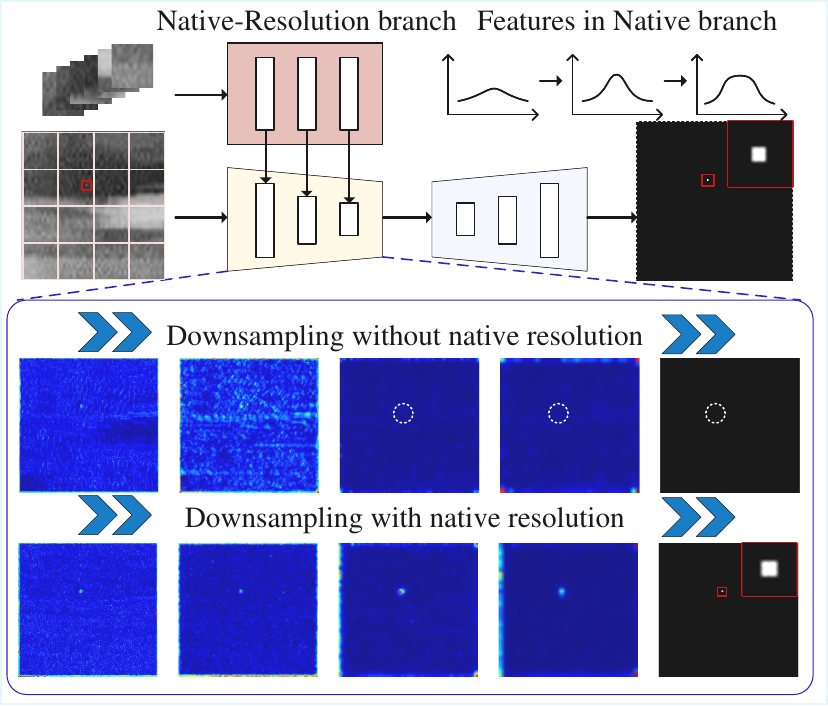}
        \caption{\rev{Visualization of feature maps across different downsampling stages. As the resolution decreases, small targets become indistinct and eventually vanish. In the native-resolution branch, the responses of small targets are progressively enhanced. By introducing this native-resolution branch, even extremely small targets can be preserved and successfully detected.}}
    \label{fig:downsampling}
    \vspace{-3mm}
\end{figure}

Unlike generic object detection tasks, \zc{which often deal with larger objects that exhibit rich textures and clear semantic boundaries,} \zc{IRSTD presents a unique challenge. It focuses on sparse, small (usually less than 32$\times$32 pixels~\cite{10914515}), textureless targets that lack meaningful structural cues. These targets emit limited infrared radiation, appearing as faint, isolated spots. This makes them highly vulnerable to being overwhelmed by environmental noise or complex backgrounds like clouds, ocean waves, or urban structures, posing significant difficulties for reliable detection in real-world settings.}


Various IRSTD methods have been proposed to address these challenges. Early approaches primarily relied on handcrafted priors, which performed well in simple scenarios but struggled in more complex environments where background noise and interference significantly obscured the targets, resulting in suboptimal performance, particularly in dynamic or cluttered scenes~\cite{han2020infrared, zhang2019infrared,9208789}. With the rise of deep learning, CNN-based methods became dominant due to their ability to automatically learn hierarchical features from input data through stacked convolutional layers~\cite{ronneberger2015unet,zhang2022isnet,huang2025Dstransnet,ma2023msma}. However, these methods are limited in capturing long-range dependencies and often struggle to model the global background.
Recent studies have addressed these limitations by incorporating Vision Transformer (ViT) and hybrid CNN-ViT architectures. ViT's self-attention mechanism improves the understanding of distant relationships within the image, while hybrid architectures combine CNNs for efficient local feature extraction and ViTs for enhanced global context awareness. This combination boosts detection capability, particularly in complex, cluttered, and low-contrast environments~\cite{zhang2024irsam,9771052,zhu2024toward,chen2026dcganet,ma2024mcdnet }.

\begin{table}[t!]
\centering
\caption{
Detection performance with native-resolution and 2$\times$ downsampling.
Values are reported as {IoU / $P_d$ / $F_a$}.
}
\label{tab:downsampling_ablation}
\resizebox{\columnwidth}{!}{
\begin{tabular}{lcc}
\toprule
{Dataset} & {Native} & {2$\times$ Downsampling} \\
\midrule
IRSTD-1k     & 71.76 / 93.11 / 8.97     & 64.83 / 89.76 / 14.79 \\
NUDT-SIRST   & 93.88 / 98.60 / 4.71     & 80.92 / 92.77 / 12.82 \\
MDFA         & 41.24 / 87.50 / 73.93    & 38.63 / 81.34 / 184.45 \\
SIRSTAUG     & 72.76 / 99.17 / 34.66    & 67.62 / 94.15 / 79.28 \\
\bottomrule
\end{tabular}
}
\end{table}


Despite significant progress achieved by deep learning-based methods, existing approaches are structurally adapted from general-purpose vision models, such as classification or semantic segmentation backbones, which typically adopt a hierarchical structure with iterative spatial downsampling. While this adoption streamlines development and delivers commendable performance, the vast size difference between small targets and objects in general recognition tasks raises a critical question: does frequent downsampling compromise the detection of small targets by discarding crucial target details?

To validate this concern, we visualized feature maps at various stages of a typical downsampling-based network~\cite{zhang2024unleashing,zhu2025toward}. As shown in Figure~\ref{fig:downsampling}, small infrared targets are progressively diminishing and disappearing after repeated resolution reduction. To quantify this effect, we train the model on several datasets under two configurations: native resolution and 2$\times$ downsampling. As shown in Table~\ref{tab:downsampling_ablation}, performance drops in the downsampled setting across all datasets, confirming that downsampling leads to the loss of crucial target details for small target detection.
These observations suggest that the conventional hierarchical design, 
originally developed for natural image understanding, 
may not be well aligned with the intrinsic characteristics of infrared small target detection. 
Unlike general object detection tasks where objects occupy sufficient spatial support 
for semantic abstraction, infrared small targets typically consist of only a few pixels 
and lack rich structural cues. 
Once subjected to stride-4 or stride-8 downsampling, 
their responses may collapse into background noise, 
making subsequent feature modeling unreliable. 
Therefore, for IRSTD, preserving native-resolution information is not merely a design preference, 
but a structural necessity aimed at preventing signal-to-noise structural collapse.

\zc{However, directly processing native-resolution data presents significant \zqm{technical} challenges\cite{10534872}. Direct native-resolution processing is computationally prohibitive, as pixel-level modeling scales quadratically with spatial resolution. Moreover, in the context of small-target detection, the primary challenge lies in the fact that only a tiny fraction of the image contains relevant target information. For example, in IRSTD tasks, small targets may occupy less than 1$\%$ of the total image area. This redundancy not only leads to an inefficient use of computational resources but also exacerbates the difficulty of distinguishing meaningful features from noise. Therefore, the real challenge lies in effectively reducing this redundancy without losing essential target cues. Achieving both efficient processing and accurate detection becomes virtually impossible without effective strategies to minimize  the overwhelming redundancy of high-resolution processing while still enabling the model to capture and preserve fine details critical for accurate detection.
}

\zc{To address this challenge, we propose Na-IRSTD, a novel framework that fully leverages the advantages of native-resolution representations for infrared small target detection. To handle the complexity of high-dimensional data, we streamline a comprehensive pipeline encompassing the model design, token reduction strategy, and multi-stage training. Specifically, Na-IRSTD introduces an effective native-resolution information extractor that detects target regions and provides fine-grained details to the model, significantly enhancing its ability to localize small, low-contrast targets that would otherwise be lost during downsampling. To further optimize this process, we propose a token reduction strategy tailored to small targets, efficiently filtering out redundant native-resolution tokens while preserving key details. Additionally, we introduce a two-stage training strategy that decouples the token reduction and infrared object detection tasks, thus reducing training complexity by addressing these two tasks separately. Finally, we curate the IRSTD-Hard dataset, which consists exclusively of targets smaller than 20 pixels. This dataset more accurately reflects the challenge of detecting tiny infrared targets and aims to advance the development of more effective detection methods. Extensive experiments demonstrate that Na-IRSTD achieves state-of-the-art performance.}

The main contributions of this work are summarized as follows:
\begin{itemize}
   \rev{\item {We propose Na-IRSTD, an efficient framework for infrared small target detection that preserves native-resolution evidence before irreversible downsampling, thereby retaining subtle target cues while balancing high detection accuracy and computational cost.}
    \item {We design a patch-based native-resolution branch composed of Patchwise Detail Extraction (PDE) and Global Patch Mixer (GPM), which preserves local infrared details while modeling inter-patch dependencies at manageable computational cost.}
    \item {We introduce a two-stage training strategy to reduce training complexity and implement token reduction. This approach effectively filters out redundant native-resolution tokens while retaining critical details.}
    \item We curate IRSTD-Hard, a benchmark for evaluating infrared small target detection in the extreme small-object regime. Our experimental results on IRSTD-Hard and other representative datasets indicate that Na-IRSTD alleviates the information-loss issue associated with downsampling and achieves consistent improvements over strong baselines. }
\end{itemize}

The structure of this manuscript is outlined as follows: Section II reviews current approaches relevant to our method, Section III introduces the Na-IRSTD technique, Section IV presents the experimental results and analysis on several  datasets, and Section V concludes with a summary of the findings.
\section{Related Work}

\subsection{IRSTD Methods}

Existing infrared small target detection methods can be broadly categorized into traditional and deep learning-based approaches. Traditional methods, such as IPI~\cite{gao2013infrared}, NRAM~\cite{zhang2018infrared}, RIPT~\cite{dai2017reweighted}, PSTNN~\cite{zhang2019infrared}, and MSLSTIPT~\cite{sun2020infrared}, are typically model-driven and rely heavily on handcrafted priors such as filters, human visual system models, or sparse representations. These methods are designed to capture certain characteristics of infrared small targets, such as local contrast, edges, or gradients, and can work effectively in relatively controlled environments. However, their performance often degrades in complex scenes, where strong background clutter, varying illumination, and target appearance changes obscure the weak target signals.

\rev{With the availability of larger datasets and the rapid development of deep learning, learning-based IRSTD methods have achieved substantial progress. CNN-based methods improve target representation by progressively extracting hierarchical features and suppressing background interference. For example, ACMNet~\cite{dai2021asymmetric} enhances target features through asymmetric contextual modulation, while DNANet~\cite{li2022dense} introduces dense nested attention to alleviate feature loss during propagation. More recent methods further strengthen global modeling and long-range dependency learning by incorporating Transformers or hybrid architectures. RKformer~\cite{zhang2022rkformer} combines CNNs and Transformers with a Runge--Kutta design, Dim2Clear~\cite{zhang2024single} enhances visibility via spatial-frequency interaction, FAA-Net~\cite{zhuang2025faa} introduces frequency-aware attention, and MiM-ISTD~\cite{chen2024mim}, IRMamba~\cite{zhang2025irmamba}, MOE-IR~\cite{weng2025moe}, and SAIST~\cite{zhang2025saist} explore Mamba-based modeling, expert specialization, or multimodal guidance to improve robustness in cluttered scenes.

Recent representative studies have further explored several directions for IRSTD, such as frequency-domain feature enhancement~\cite{zhu2025toward,hu2026dual}, global--local or contextual feature fusion~\cite{ma2025temporal,ma2024mcdnet,zhu2024toward,zhu2025shifting,chen2026dcganet}, and high-resolution representation learning or detail recovery~\cite{ma2023msma,huang2025Dstransnet,fan2024diffusion}. These studies further confirm the importance of preserving or enhancing subtle target cues for IRSTD.

Nevertheless, most existing IRSTD methods still inherit the hierarchical encoding paradigm from general-purpose vision models, where feature extraction is coupled with repeated spatial downsampling. As a result, the subtle responses of extremely small targets may already be weakened or even lost before later enhancement modules are applied. Methods based on feature enhancement, global--local interaction, or high-resolution reconstruction can improve degraded representations, but they still mainly operate on dense features after or during hierarchical encoding. In contrast, the key question addressed in this work is how to preserve native-resolution evidence before irreversible information loss occurs, while doing so in a sparse and target-aware manner suitable for the extreme sparsity of IRSTD.}
\subsection{Native-Resolution Modeling}

To mitigate the loss of spatial precision caused by downsampling, a growing body of work has explored direct modeling at native or near-native resolution. HRNet~\cite{sun2019hrnet} maintains high-resolution representations throughout the network by connecting multi-scale branches in parallel and repeatedly exchanging information, enabling accurate localization with strong semantics.  However, this comes at the cost of significant computational overhead due to its multi-resolution feature extraction and fusion.
Dilated convolutions can expand receptive fields without reducing spatial resolution, facilitating large-context modeling at native scale. SPDConv~\cite{sunkara2022spdconv} mitigates information loss by replacing strided convolutions and pooling with a space-to-depth convolutional building block, thereby preserving fine-grained spatial information for small objects. Meanwhile, HCMamba~\cite{xu2024hc} integrates dilated convolutions with depthwise separable convolutions, significantly reducing computational costs while maintaining spatial fidelity in high-resolution medical image segmentation tasks.
CAD~\cite{hesse2023content} dynamically adjusts the resolution of input features based on their content, effectively balancing computation and accuracy. PNA~\cite{liu2023progressive} progressively aggregates neighborhood information to refine semantic segmentation predictions.
FADC~\cite{chen2024frequency} employs frequency-adaptive dilated convolutions, adjusting the receptive field based on the frequency content of the image. This approach ensures that high-frequency details, especially those in small targets, are preserved while optimizing computational efficiency.

Despite these advances, native-resolution dense processing remains computationally expensive and often redundant for sparse target scenarios, motivating the need for more selective and efficient strategies that retain high-resolution advantages while reducing cost.
\subsection{Token Reduction Strategies}

To alleviate the high computational cost associated with processing dense high-resolution data, recent studies have proposed various token reduction methods that adaptively prune, halt, or merge redundant tokens while preserving essential information, thereby improving computational efficiency. Merging-based methods, such as Token Merging (ToMe)~\cite{bolya2023tome}, reduce the token count by merging similar tokens according to token similarity, which lowers computational cost while retaining most informative content. Pruning- and halting-based methods, such as DynamicViT~\cite{rao2021dynamicvit} and A-ViT~\cite{yin2022avit}, learn to remove or halt less informative tokens according to learned importance or halting scores. ATF~\cite{naruko2025speed} further introduces attention-aware token filtering to accelerate Vision Transformer models by filtering tokens before they are fed into the transformer encoder.
Adaptive computation methods, such as AdaViT~\cite{meng2022adavit} and Evo-ViT~\cite{xu2022evovit}, exploit input-dependent redundancy across network layers. AdaViT learns instance-specific usage policies for patches, attention heads, and transformer blocks, while Evo-ViT updates informative and less informative tokens through different computation paths, maintaining spatial structure and information flow. For dense prediction tasks, methods such as DToP~\cite{tang2023dtopp} and DoViT~\cite{liu2024dovit} are specifically designed to reduce computation while preserving dense spatial predictions. DToP finalizes predictions for easy tokens at intermediate layers and keeps representative context tokens, whereas DoViT dynamically stops easy tokens from self-attention computation and reconstructs token positions for semantic mask prediction. These methods demonstrate that token reduction must be task-aware, especially when spatial accuracy is required.

\rev{However, these strategies are primarily developed for natural image understanding, where informative content is relatively dense and token importance can often be estimated from generic similarity or saliency cues. In IRSTD, by contrast, targets are extremely small, sparse, and easily confused with background clutter. Under this setting, similarity-based pruning or merging may suppress the very weak target cues that need to be preserved. Therefore, token reduction for IRSTD should not be treated as a generic efficiency problem alone. Balancing computational efficiency and the preservation of small target cues remains a challenge.}

\section{Methodology}

\begin{figure*}[t!]
    \centering
    \includegraphics[width=1\linewidth]{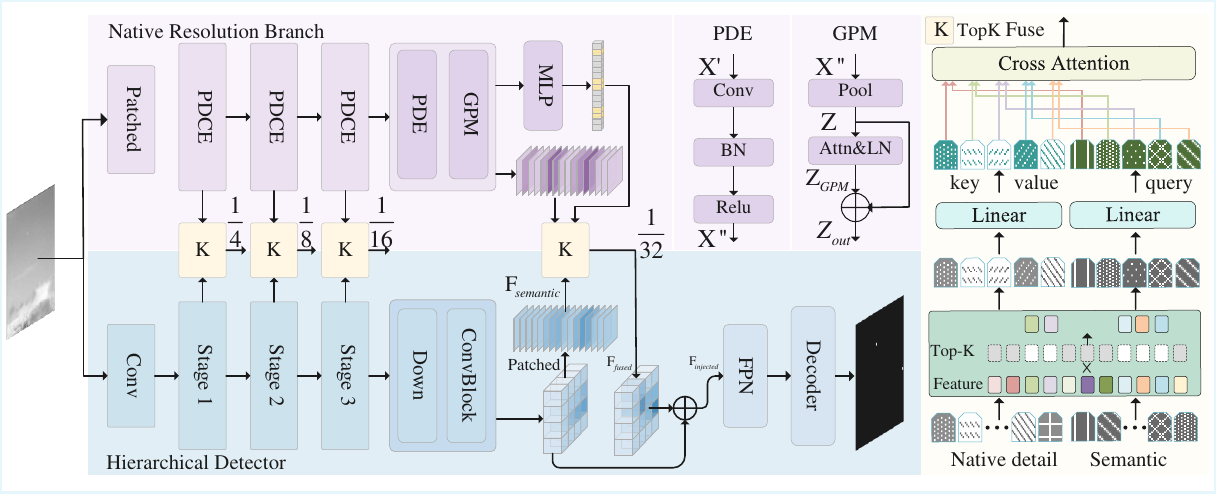}
        \caption{\rev{Overview of the proposed Na-IRSTD architecture. The upper pathway is the native-resolution branch and the lower pathway is the hierarchical backbone encoder. PDCE denotes the Patchwise Detail-Context Encoder, composed of PDE and GPM. PDE denotes Patchwise Detail Extraction, and GPM denotes Global Patch Mixer. For clarity, only one representative stage is expanded to illustrate the detailed operations, while the remaining stages follow the same design pattern.}}
    \label{fig:architecture}
\end{figure*}
\subsection{Overview of Na-IRSTD Framework}

To address the \zqm{information loss issue of existing IRSTD methods}, we develop Na-IRSTD—an end-to-end framework that encompasses model design, a tailored training pipeline, an efficient token reduction strategy, and a challenging benchmark. This comprehensive suite facilitates a more effective and systematic exploration of native-resolution processing for IRSTD.

As shown in Figure~\ref{fig:architecture}, the overall architecture of Na-IRSTD comprises two complementary pathways: a DEtection TRansformer (DETR)-like hierarchical detector $\mathcal{E}(\cdot)$~\cite{zhang2024unleashing} and a native-resolution branch $\mathcal{N}(\cdot)$.

Given an input image \(I \in \mathbb{R}^{3 \times H\times W}\), the backbone encoder \(\mathcal{E}(\cdot)\) extracts a hierarchy of semantic features from the entire image. In parallel, the native-resolution branch \(\mathcal{N}(\cdot)\)  is introduced to extract the native-resolution information, which is typically lost in conventional encoders. Before being processed by this branch, the input image is divided into non-overlapping patches of size \(P\times P\), resulting in \(N = \frac{H}{P} \times \frac{W}{P}\) patches. The extracted feature representations from both branches are described as follows:
\begin{align}
F_s^{(l)} &= \mathcal{E}(I) \in \mathbb{R}^{C_l \times H_l \times W_l}, \quad l \in \{1, 2, \dots, L\}, \\
F_n^{(l)} &= \mathcal{N}(I) \in \mathbb{R}^{N \times C_l \times P \times P}, \quad l \in \{1, 2, \dots, L\},
\end{align}
where \( l \in \{1, 2, \dots, L\} \) denotes the network stage.

To alleviate the excessive computation burden and redundancy introduced by dense patch processing, we propose an efficient token reduction strategy that enables the model to selectively retain the most informative regions. Relevance scores $\mathcal{S}=\{s_1, \dots, s_N\} \in \mathbb{R}^N$ are learned for each patch in $F_n^{(L)}$. \zqm{The top-K patches for each stage are selected as:}
\begin{equation}
\mathcal{K}^{(l)} = \mathrm{Top}\text{-}K(F_n^{(l)}, \mathcal{S}) \in \mathbb{R}^{K \times C_l \times P \times P},
\end{equation}
where $C_l$ denotes the feature dimension at level $l$, and $K \ll N$ ensures computational efficiency.

Then, the cross-scale fusion module $\mathcal{F}(\cdot)$ projects the selected patches $\mathcal{K}^{(l)}$ into the semantic features $F_s^{(l)}$, producing a fused feature map $F^{(l)} \in \mathbb{R}^{C_l \times H_l \times W_l}$ that integrates both global context and fine-grained details:
\begin{equation}
F^{(l)} = \mathcal{F}(F_s^{(l)}, \mathcal{K}^{(l)}).
\end{equation}

Finally, the fused features from all layers \(F = \{F^{(1)}, F^{(2)}, \dots, F^{(L)}\}\) are passed to the decoder \(\mathcal{D}(\cdot)\), which generates the final prediction.

\subsection{\zc{Taming Feature Learning in Native Resolution}}

\subsubsection{Details of the Native-Resolution Branch} \label{Details of the Native-Resolution Branch}
Typically, detection and segmentation-oriented Vision Transformers (ViTs)~\cite{Kirillov_2023_ICCV} address the challenge of high computational cost by downsampling input features before feeding them into subsequent layers. This design is necessary due to the quadratic complexity of the self-attention mechanism, which scales poorly with input resolution and quickly becomes computationally prohibitive.

\rev{To enable feature learning at native resolution, we introduce a native-resolution branch that directly operates on native-resolution patches. Unlike a standard Vision Transformer (ViT), which embeds the whole image into flattened spatial tokens for global self-attention, our branch adopts a 2D+1D structure inspired by video modeling. The branch is built upon a Patchwise Detail-Context Encoder (PDCE), which consists of Patchwise Detail Extraction (PDE) and a Global Patch Mixer (GPM). It first performs convolutional detail extraction within each patch and then applies self-attention over pooled patch tokens. This design preserves local infrared details while efficiently modeling inter-patch dependencies.}

Specifically, each input image is divided into $N$ non-overlapping patches of size $P \times P$, forming a 5D pseudo-video tensor $\mathbf{X} \in \mathbb{R}^{B \times N \times C \times P \times P}$, where $B$ is the batch size, $C$ is the input channel dimension, and $N = \frac{H}{P} \times \frac{W}{P}$ is the total number of patches. To encode local spatial features, we first reshape $\mathbf{X}$ into a 4D tensor:
\begin{equation}
\mathbf{X}' = \mathrm{Reshape}(\mathbf{X}) \in \mathbb{R}^{(B \cdot N) \times C \times P \times P},
\end{equation}
and then apply a shared convolutional encoder $\mathrm{PDE}(\cdot)$ (patchwise detail extractor, see Figure~\ref{fig:architecture}) to process each patch independently:
\begin{equation}
\mathbf{X}'' = \mathrm{PDE}(\mathbf{X}') \in \mathbb{R}^{(B \cdot N) \times C' \times P \times P},
\end{equation}
where $C'$ denotes the output channel dimension. The resulting features are then reshaped back to a 5D tensor to preserve the pseudo-frame structure.

To capture long-range dependencies across patches, we introduce the \emph{Global Patch Mixer} (GPM, Figure~\ref{fig:architecture}), which aggregates global context from the sequence of patch-level features. To enable efficient global modeling, we reduce the spatial dimension of each patch by applying average pooling to obtain a compact token representation:
\begin{equation}
\mathbf{z}_i = \mathrm{AvgPool}(\mathbf{X}''_i),
\end{equation}
resulting in a sequence of patch-level tokens $\mathbf{Z} = [\mathbf{z}_1, \mathbf{z}_2, \ldots, \mathbf{z}_N] \in \mathbb{R}^{B \times N \times C'}$. 
Given the sparse nature of infrared targets—where only a small fraction of patches contain meaningful information—this patch-level abstraction is sufficient to capture long-range dependencies without incurring the computational cost of pixel-level attention.

Unlike conventional attention in ViTs that operates over flattened spatial tokens ($H \times W$), GPM applies self-attention across a sequence of pooled patch tokens $\mathbf{Z} \in \mathbb{R}^{B \times N \times C'}$, 
where $N = \frac{H \times W}{P^2}$. This inter-token attention is defined as:
\begin{equation}
\mathbf{Z}_{\text{GPM}} = \mathrm{Softmax}\left(\frac{\mathbf{Q} \mathbf{K}^T}{\sqrt{d_k}}\right) \mathbf{V},
\end{equation}
with $\mathbf{Q}, \mathbf{K}, \mathbf{V} \in \mathbb{R}^{B \times N \times d_k}$. 
This formulation models global dependencies across spatial regions while avoiding the quadratic complexity in $H \times W$ typical of pixel-level attention. The output is then fused with the original features via residual addition:
\begin{equation}
\mathbf{Z}_{\text{out}} = \mathbf{Z} + \mathbf{Z}_{\text{GPM}}.
\end{equation}
This fusion enables each patch to benefit from globally aggregated cues while maintaining its local characteristics.



By decoupling local detail extraction and global context modeling, our framework enables efficient and expressive feature learning at native resolution, preserving fine-grained spatial information. Convolutional encoding operates within each patch to capture local details, while GPM efficiently captures long-range dependencies across patch tokens, allowing the model to effectively integrate both local and global context.

This design avoids the downsampling caused by the initial pooling layers in ViTs, thereby preserving the original resolution features. Furthermore, it reduces the computational burden of pixel-level ViTs, lowering the complexity from $O((H \times W)^2)$ to $O\left(\left(\frac{H \times W}{P^2}\right)^2\right)$
offering a scalable, resolution-preserving alternative to both ViTs and CNNs.
\subsubsection{Token Selection via Relevance Estimation}
Although the native-resolution branch preserves detailed spatial information, it inevitably introduces considerable redundancy, as infrared targets typically occupy only a small subset of image patches. To enhance efficiency and direct model capacity toward informative regions, we introduce a lightweight scoring mechanism that enables the model to learn which patches are likely to contain targets. Unlike existing token pruning or merging approaches designed for natural image understanding, which typically rely on token similarity to perform reduction, our method is tailored for the unique challenges posed by IRSTD. 
In IRSTD, targets are extremely small, sparse, and often indistinguishable from the background without fine-grained resolution. Similarity-based token fusion strategies in this context risk discarding target-relevant tokens or severely degrading spatial resolution, both of which lead to significant performance drops.
Considering the sparseness of infrared targets, Na-IRSTD reformulates the token selection problem as a patch-wise classification task, where the objective is to score the probability that each image patch contains a potential target. Specifically, an MLP \zqm{layer} is appended to the output of the native-resolution branch to predict a scalar relevance score for each patch, reflecting its likelihood of being target-relevant. To supervise the scoring network more effectively, we adopt a soft-labeling strategy. We assign labels based on the proximity to target centers, allowing for more nuanced supervision that enhances the model’s ability to focus on relevant target areas. For each image, let $\mathcal{M} \in \{0,1\}^{H \times W}$ denote the binary ground-truth mask, and $\mathcal{T} = \{(x_i, y_i)\}_{i=1}^{N}$ be the set of target centers. We define the soft target weight of a pixel at location $(x,y)$ as: 
\begin{equation} 
w(x, y) = \exp\left( -\frac{1}{2\sigma^2} \cdot \min_{(x_i, y_i) \in \mathcal{T}} \left[(x - x_i)^2 + (y - y_i)^2 \right] \right), 
\label{eq:gaussian_label} 
\end{equation} 
where $\sigma$ is a hyperparameter controlling the spatial decay rate. For each image patch $\mathcal{P}_j$, the soft label is computed as the average of weights over all pixels within the patch: \begin{equation} 
\tilde{y}_j = \frac{1}{|\mathcal{P}_j|} \sum_{(x,y) \in \mathcal{P}_j} w(x, y). 
\end{equation} 
This formulation produces continuous-valued labels in $[0,1]$, which serve as patch-level ground-truth scores. Compared to hard labels, this soft supervision smooths the semantic boundary between foreground and background, mitigates class imbalance, and provides more informative gradients during training. The scoring network is built upon the feature extraction module described in Section \ref{Details of the Native-Resolution Branch}. Specifically, after the patch-wise detail-to-semantic feature extraction, a lightweight multi-layer perceptron (MLP) is appended to predict a scalar score for each patch, indicating the likelihood of target presence. 
The scoring network is trained using Binary Cross-Entropy (BCE) loss between the predicted score $s_j$ and the soft label $\tilde{y}_j$:
\begin{equation}
{L}_{\text{score}} = \frac{1}{N} \sum_{j=1}^N \mathrm{BCE}(s_j, \tilde{y}_j).
\label{score}
\end{equation}
Compared to hard binary labels, this soft-labeling approach is particularly beneficial in the context of extreme class imbalance inherent to IRSTD. Since positive patches are exceedingly sparse, directly assigning binary labels often leads to unstable optimization and poor generalization. In contrast, soft labels provide a distance-aware, differentiable supervision signal that facilitates more stable and effective learning.

The learned relevance scores enable selective processing of only the most informative patches.
Specifically, instead of modeling interactions over all $N$ native patches, 
we retain only the Top-$K$ patches with the highest scores for subsequent cross-scale injection.
Dense native-resolution modeling processes all $N$ patches, 
leading to a patch-level attention complexity proportional to $O(N^2)$.
With Top-$K$ selection, the effective token number is reduced to $K$, 
resulting in a complexity of $O(K^2)$.
Accordingly, the token reduction ratio is
$
1 - \frac{K}{N}.
$
Under the unified configuration ($256\times256$ input and $P=32$), 
$N=64$.
With $K=5$, this corresponds to a $92.2\%$ token reduction 
relative to dense native-resolution processing.

\subsubsection{Native-Resolution Injection and Optimization}
\label{sec:inject}

To effectively integrate native-resolution cues into hierarchical backbone representations, we design a cross-scale injection mechanism together with a two-stage optimization strategy. This design addresses both spatial-semantic misalignment and training instability caused by extreme foreground sparsity.

\textbf{Cross-scale injection.}
\rev{Given an input of size $H\times W$ and a non-overlapping patch size $P\times P$, the image is partitioned into
$N=\frac{H}{P}\times\frac{W}{P}$ patches. For each stage $l$, we select a Top-$K$ index set
$\mathcal{J}^{(l)}=\{j_1,\ldots,j_K\}$ according to the learned relevance scores.
For the backbone feature map $F_s^{(l)}\in\mathbb{R}^{B\times C_l\times H_l\times W_l}$ with downsampling factor $s_l$,
we set $P'_l=P/s_l$ and partition $F_s^{(l)}$ into $N$ non-overlapping windows of size $P'_l\times P'_l$, producing a patch-aligned semantic token set
$\Phi^{(l)}\in\mathbb{R}^{B\times N\times C_l\times P'_l\times P'_l}$.
For a selected patch index $j$, let $u_j=\left\lfloor \frac{j}{W/P} \right\rfloor$ and $v_j=j-u_j(W/P)$ denote its row and column coordinates on the original patch lattice. The corresponding semantic window at stage $l$ therefore starts from $(r_j^{(l)},c_j^{(l)})=(u_jP'_l,v_jP'_l)$, so the aligned semantic regions can be obtained by indexing $\Phi^{(l)}$ with $\mathcal{J}^{(l)}$ along the patch dimension. In this way, each selected native patch shares the same lattice index as its semantic window, establishing a deterministic patch-to-window correspondence without resizing or interpolation.
Let the Top-$K$ selected patches at stage $l$ be 
$\mathcal{K}^{(l)} \in \mathbb{R}^{B \times K \times D_l \times P \times P}$,
and the aligned semantic regions be 
$F_{\text{sem}}^{(l)} \in \mathbb{R}^{B \times K \times C_l \times P'_l \times P'_l}$.
A stage-specific $1\times1$ projection is applied to match channel dimensions from $D_l$ to $C_l$.
Then we perform cross-attention between the aligned semantic regions and the selected native-resolution patches:
\begin{equation}
F_{\text{inj}}^{(l)} 
= F_{\text{sem}}^{(l)} 
+ \alpha_l \cdot 
\mathrm{CrossAttn}\!\left(
F_{\text{sem}}^{(l)}, 
\mathcal{K}^{(l)}
\right),
\label{eq:cross_injection}
\end{equation}
where $\alpha_l$ is a learnable scaling factor. Finally, the updated regions $F_{\text{inj}}^{(l)}$ are written back to $F_s^{(l)}$ at the same indices $\mathcal{J}^{(l)}$,
yielding fused multi-scale representations for decoding.}

\textbf{Two-stage optimization.}
To improve convergence stability, we decouple patch relevance learning from dense segmentation.

In Stage 1, only the native-resolution branch $\mathcal{N}(\cdot)$ and the scoring MLP are trained using the soft supervision in Eq.~(\ref{eq:gaussian_label}) and the BCE loss in Eq.~(\ref{score}), while the backbone, fusion module, and decoder remain inactive.

In Stage 2, we initialize the native-resolution branch and the scoring MLP with the pretrained weights from Stage 1. The scoring MLP is then kept fixed, while the native-resolution branch, backbone encoder, fusion module, and decoder are optimized using the segmentation loss
\begin{equation}
L_{\text{seg}} = L_{\text{BCE}} + L_{\text{Dice}}.
\end{equation}
where BCE loss is defined as:
\begin{equation}
   L_{\text{BCE}} = - \sum_{i=1}^{N} [y_i \log(p_i) + (1 - y_i) \log(1 - p_i)], 
\end{equation}
where $y_i$ is the true label for the $i$-th sample and $p_i$ is the predicted probability for the $i$-th sample being in the positive class.  Dice Loss is defined as:
\begin{equation}
    L_{\text{Dice}} = 1 - \frac{2 | \hat{y} \cap y |}{|\hat{y}| + |y|},
\end{equation}
where \( \hat{y} \) is the predicted segmentation mask, and \( y \) is the ground truth segmentation mask. The BCE loss ensures the binary classification of each pixel (target or background), while the Dice Loss improves the overlap between predicted and true target regions.

In Stage 2, the fixed scoring MLP predicts patch relevance scores for native-resolution patches, and deterministic hard Top-$K$ selection is applied according to the predicted scores. Since hard Top-$K$ selection is non-differentiable with respect to the patch scores, gradients from the segmentation loss are not propagated through the selection operation to the scoring MLP. The same deterministic hard Top-$K$ strategy is used during both training and inference.

\begin{algorithm}[t]
\caption{Two-stage training of Na-IRSTD}
\label{alg:two_stage}
\begin{algorithmic}[1]
\REQUIRE Training images $\{I\}$, masks $\{\mathcal{M}\}$

\STATE \textbf{Stage 1: Patch relevance pretraining}
\STATE Initialize $\mathcal{N}$ and scoring MLP; freeze $\mathcal{E},\mathcal{F},\mathcal{D}$
\FOR{each minibatch}
    \STATE Compute patch scores $\mathbf{s}$ and soft labels $\tilde{\mathbf{y}}$
    \STATE Update $\mathcal{N}$ and scoring MLP via $L_{\text{score}}$
\ENDFOR

\STATE \textbf{Stage 2: Segmentation fine-tuning}
\STATE Load pretrained $\mathcal{N}$ and scoring MLP; freeze scoring MLP
\FOR{each minibatch}
    \STATE Predict scores $\mathbf{s}$ using the fixed scoring MLP
    \STATE Select the Top-$K$ patches according to $\mathbf{s}$
    \STATE Perform cross-scale fusion and predict mask
    \STATE Update $\mathcal{N}, \mathcal{E}, \mathcal{F}, \mathcal{D}$ via $L_{\text{seg}}$
\ENDFOR

\STATE \textbf{Inference:} deterministic hard Top-$K$ selection using predicted scores

\end{algorithmic}
\end{algorithm}

\begin{table*}[!h]
\caption{Characteristics of the {test splits} in Various Infrared Small Target Detection Datasets. }
\centering
\renewcommand\arraystretch{1.25}
\setlength{\tabcolsep}{4pt}
\begin{tabular}{lcccccc}
\toprule
{Dataset} & {Real or Simulated} & {Dataset Size} & {Image Size} & {Target Size} & {Avg. Target Size} & {Background Type} \\ \midrule
{NUDT}    & Simulated                  & 300           & 256 $\times$ 256    &  1\ensuremath{\sim}98 pixels & 31.4 pixels   & city, cloud, field  \\ 
{IRSTD-1k}   & Real                      & 200           & 256 $\times$ 256    & 1\ensuremath{\sim}329 pixels & 11.9 pixels   & sea, field, mountain \\ 
{SIRSTAUG}   & Real                      & 535           & 256 $\times$ 256    & 5\ensuremath{\sim}364 pixels & 82.8 pixels   & mountain, field, sea\\ 
{IRSTD-Hard}   & Real/Simulated                      & 294           & 256 $\times$ 256    & 1\ensuremath{\sim}20 pixels & 7.5 pixels   & sea, field, mountain, city, and cloud\\ 
 \bottomrule
\end{tabular}

\label{tab:datasets}
\end{table*}

\subsection{IRSTD-Hard Benchmark}

Most existing infrared small target detection datasets contain a considerable portion of medium-sized targets, as summarized in Table~\ref{tab:datasets}.  Statistics in Table~\ref{tab:datasets} are computed on the official test partitions, and our IRSTD-Hard benchmark is also constructed solely from test partitions. While valuable for general evaluation, they are not explicitly designed to assess performance in the extreme small-object regime. To systematically investigate this setting, we construct a benchmark named IRSTD-Hard, which focuses exclusively on targets smaller than 20 pixels.

The benchmark is constructed exclusively from the test partitions of three publicly available datasets: IRSTD-1k, NUDT-SIRST, and IRDST. It contains 106 images from IRSTD-1k, 101 from NUDT-SIRST, and 87 from IRDST, resulting in a total of 294 images. A uniform size threshold is applied across all source datasets based on the annotated target area, and no images from the corresponding training subsets are included. All images are resized to 256$\times$256 for evaluation consistency.

\begin{figure}[t]
    \centering
    \includegraphics[width=1.0\linewidth]{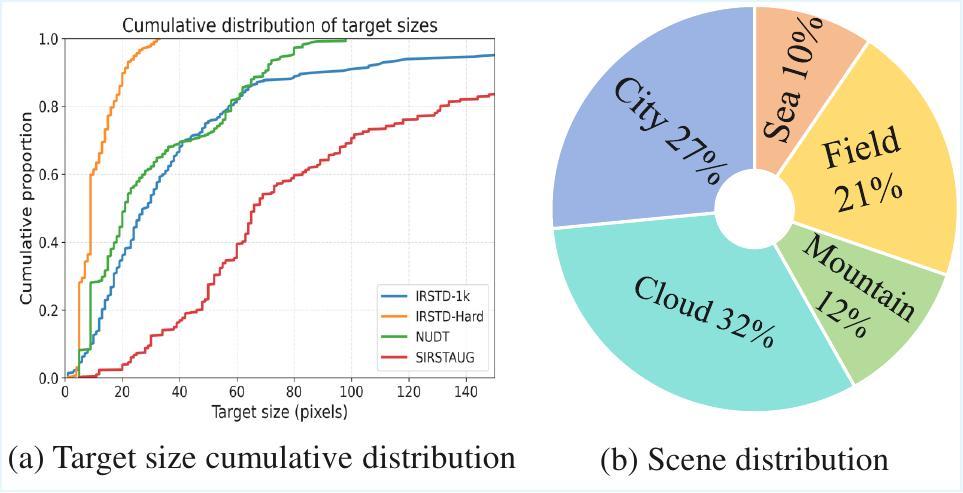}
    \caption{\rev{Statistical characteristics of IRSTD-Hard. (a) Target size cumulative distribution of different IRSTD datasets. (b) Scene distribution of IRSTD-Hard.}}
    \label{fig:hard_stats}
\end{figure}

\rev{The average target size is 7.5 pixels, and the median size is 6 pixels, indicating a distribution concentrated on extremely small objects. As shown in Fig.~\ref{fig:hard_stats}(a), the cumulative target-size curve of IRSTD-Hard rises much faster than those of the other IRSTD datasets, confirming that our benchmark contains a substantially larger proportion of extremely small targets. Among the 294 images, 73 contain multiple targets. The dataset includes both real and simulated infrared imagery covering scene categories such as sea, field, mountain, urban areas, and cloud backgrounds. Fig.~\ref{fig:hard_stats}(b) further shows that IRSTD-Hard maintains diverse scene composition rather than focusing on a single background type. The relatively balanced source distribution (106/101/87), together with the strict test-only construction, helps alleviate potential source bias.}
All models are trained on the training split of NUDT-SIRST
and evaluated on IRSTD-Hard.
This protocol guarantees strict train–test disjointness at the image level
and evaluates robustness under cross-dataset distribution shifts.

\begin{table*}
\renewcommand\arraystretch{1.25}
\centering
\caption{Comparison with other state-of-the-art methods on four datasets. The metrics considered include IoU ($10^{-2}$), $P_d$ ($10^{-2}$), and $F_a$ ($10^{-6}$).}
	\label{tab:sota}
	\begin{tabular}{l|c|ccc|ccc|ccc|ccc}
		\noalign{\hrule height 1pt}
		\multirow{2}{*}{Method} & \multirow{2}{*}{Publish} 
		& \multicolumn{3}{c|}{IRSTD-1k}   
		& \multicolumn{3}{c|}{SIRSTAUG}   
		& \multicolumn{3}{c|}{NUDT-SIRST} 
		& \multicolumn{3}{c}{IRSTD-Hard} \\ 
		&  
		& IoU $\uparrow$  & $P_d$ $\uparrow$  & $F_a$ $\downarrow$
		& IoU $\uparrow$  & $P_d$ $\uparrow$  & $F_a$ $\downarrow$
		& IoU $\uparrow$  & $P_d$ $\uparrow$  & $F_a$ $\downarrow$
		& IoU $\uparrow$  & $P_d$ $\uparrow$  & $F_a$ $\downarrow$ \\
		\noalign{\hrule height 1pt}
		NRAM \cite{zhang2018infrared}  & RS'18   & 9.88 & 72.48 & 24.73 
        & 8.97 & 71.47 & 68.98
        & 12.08 & 72.58 & 84.77
        & 5.36 & 49.17 & 243.25 \\
		PSTNN \cite{zhang2019infrared} & RS'19   & 24.57 & 71.99 & 35.26
        & 19.14 & 73.14 & 61.58
        & 27.72 & 66.13 & 44.17
        &  11.93 & 53.38 & 124.73 \\
		\hline
		ACM \cite{dai2021asymmetric}  & WACV'21 & 63.39 & 91.25 & 8.961
        & 73.84  & 97.52  & 76.35
        & 69.26  & 96.26  & 10.27  
        &  30.87 &76.04 & 96.79 \\
        
		DNANet \cite{li2022dense} & TIP'22 
        & 68.87 & 94.95 & 13.38
        & 74.31  & 97.80   & 30.07
        & 92.09  & 99.53  & 2.35 
        & 41.51& 77.06 & 55.32\\
		ISNet \cite{zhang2022isnet}  & CVPR'22 & 68.77 & 95.56 & 15.39 
        & 72.50 & 98.41 & 28.61 
        & 88.92 & 99.12 & 4.21 
        & 40.84 & 78.31 & 53.26 \\
	    UIUNet \cite{wu2023uiu} & TIP'23
        & 69.13 &   94.27 & 16.47
        & 74.24 & 98.35 & 23.13 
        & 90.17 & 99.29 & 2.39 
        &  42.76 & 79.34 &89.16 \\
        AGPCNet \cite{10024907}    & TAES'23 & 68.81 & 94.26 & 15.85 
        & 72.88    & 97.45  & 38.33  
        & 88.71    & 97.57  & 7.54  
        &  41.47 & 79.20 & 59.53 \\
		RPCANet \cite{wu2024rpcanet} & WACV'24 & 63.21 & 88.31 & 43.9 
        & 72.54 & 98.21 & 34.14 
        & 89.03 & 97.14 & 28.7 
        &  40.35 & 79.31 & 58.68 \\
		IRMamba \cite{zhang2025irmamba} & AAAI'25 & 70.04 & 95.81 & 5.92 
        & 72.63 & 97.55 & 29.64 
        & 95.18 & 99.26 & 1.31 
        &  40.66 & 76.33 & 57.52 \\
        SAIST \cite{zhang2025saist} & CVPR'25 & 72.14 & \textbf{96.18} & 4.76 & 74.02 & 98.44 & 35.14 & 95.23 & 99.28 & 1.31 &  40.50 & 79.54 & 51.25 \\
        Na-IRSTD  &   --         & \textbf{73.42} & 95.42 & \textbf{4.33} 
        & \textbf{74.39} & \textbf{99.72} & \textbf{20.83}
        & \textbf{97.45} & \textbf{100.00} & \textbf{1.22 }
        &  \textbf{47.83} & \textbf{82.18} &\textbf{33.58} \\
	\noalign{\hrule height 1pt}
	\end{tabular}
\end{table*}
\section{Experiments}
\subsection{Experimental details}
\subsubsection{Dataset}

We evaluate our proposed method on four infrared small target detection benchmarks: IRSTD-1k~\cite{zhang2022isnet}, SIRSTAUG~\cite{10024907}, NUDT-SIRST~\cite{li2022dense}, and IRSTD-Hard. The IRSTD-1k dataset consists of 1,000 real infrared images, each containing one or more small targets. SIRSTAUG contains 8,525 images and is an augmented version of the SIRST dataset. NUDT-SIRST contains 1,327 synthetically generated infrared images with single or multiple small objects. All images are resized to $256 \times 256$ to ensure consistency. For all datasets, we split the images into 50\% training, 30\% validation, and 20\% testing subsets to ensure stable training and fair evaluation. IRSTD-Hard is used exclusively as a test set. For fairness, the models in the comparison are trained on NUDT and tested on IRSTD-Hard to evaluate their ability to detect small targets and their generalization capability.
\subsubsection{Implementation Details}
\rev{We adopt the Adam optimizer and a cosine decay learning rate scheduler. All models are trained with a batch size of 8 on dual Nvidia GeForce RTX 4090 GPUs. For the proposed token reduction, the patch size is set to $P=32$, and we keep Top-$K$ tokens per layer with $K=5$. 
\rev{Na-IRSTD is optimized in two stages. In Stage 1, only the native-resolution branch and the scoring MLP are instantiated and trained for 500 epochs with a learning rate of $1\times10^{-4}$. In Stage 2, the full model is initialized with the pretrained weights from Stage 1, and the scoring MLP is frozen. The native-resolution branch, backbone encoder, fusion module, and decoder are then trained for 1500 epochs, where the learning rate of the backbone encoder and the fusion module is set to $1\times10^{-4}$, while the native-resolution branch uses a smaller learning rate of $1\times10^{-6}$. The cosine schedule is restarted at the beginning of this stage, and deterministic hard Top-$K$ selection is applied based on the predicted relevance scores.}
For comparison, we include traditional methods, such as NRAM~\cite{zhang2018infrared} and PSTNN~\cite{zhang2019infrared}; and deep learning-based methods, such as ACM~\cite{dai2021asymmetric}, DNANet~\cite{li2022dense}, ISNet~\cite{zhang2022isnet}, UIUNet~\cite{wu2023uiu}, RPCANet~\cite{wu2024rpcanet}, IRMamba~\cite{zhang2025irmamba}, and SAIST~\cite{zhang2025saist}.}

We adopt {IoU}, {detection probability ($P_d$)}, and {false alarm rate ($F_a$)} as default metrics.
Among them, $P_d$ and $F_a$ are object-level metrics, while IoU is a pixel-level metric. Their definitions are as follows:
\begin{equation}
        P_d = \frac{TP}{TP + FN},
\end{equation}
where $TP$ and $FN$ denote the number of true positives (correctly detected targets) and false negatives (missed targets), respectively.
\begin{equation}
        F_a = \frac{FP}{N},
\end{equation}
where $FP$ is the number of false positives (unmatched detections), and $N$ is the total number of test images. 

Intersection over Union (IoU):
\begin{equation}
IoU =\frac{TP}{T + P - TP},
\end{equation}
where $T$, $P$, $TP$, and $FP$ denote the number of ground-truth positive pixels, predicted positive pixels, true positive pixels, and false positive pixels, respectively.

\begin{figure*}[t!]
    \centering
    \includegraphics[width=0.95\linewidth]{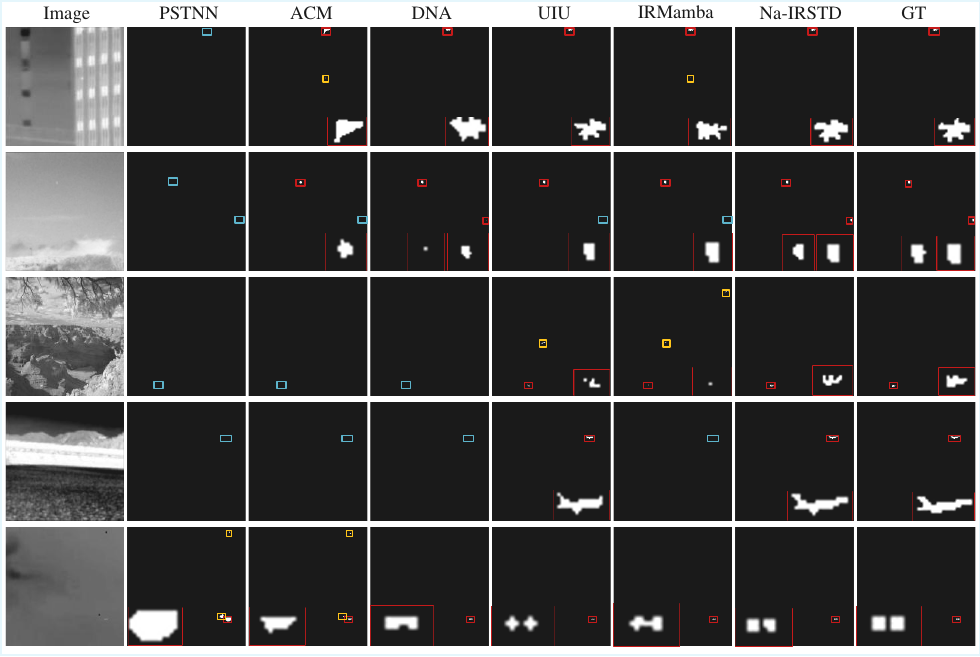}
        \caption{Visual results of different IRSTD methods. The boxes in {red}, {yellow}, and {blue} represent \textcolor{red}{correct}, \textcolor{yellow}{false}, and \textcolor{blue}{missed} detections, respectively. Close-up views are shown in the corners.}
    \label{fig:visual}
\end{figure*}

\subsection{Main Results}

\paragraph{Quantitative Results}  

As presented in Table~\ref{tab:sota}, traditional methods often struggle in complex scenarios. Methods based on general segmentation frameworks are limited in performance due to the loss of information during downsampling. By incorporating native-resolution information, our method achieves improved performance across most evaluated metrics, with particularly strong results on NUDT.
Additionally, the results on IRSTD-Hard suggest that our method generalizes well to more challenging small-target scenarios, which may partly benefit from the proposed two-stage training strategy. \rev{On IRSTD-1k, although Na-IRSTD does not achieve the highest $P_d$, its $P_d$ remains highly competitive, while the method still attains the best IoU and the lowest $F_a$. This behavior may be attributed to the fact that Na-IRSTD, trained with segmentation losses and relevance-guided sparse native-resolution fusion, tends to produce cleaner and more compact responses by focusing on the most informative target evidence and suppressing spurious background activations. Such predictions improve overlap quality and false-alarm suppression, but may be slightly less aggressive in object-level hit counting than some competing methods. Therefore, the slightly lower $P_d$ on IRSTD-1k should be interpreted together with the simultaneously improved IoU and $F_a$, indicating better overall detection quality.}

\paragraph{Visual Results} 
Figure~\ref{fig:visual} presents detection results using our method compared to other IRSTD methods. These visual comparisons are consistent with the quantitative results and suggest that the integration of native-resolution features is beneficial in cluttered scenes. These results highlight our method’s ability to effectively capture fine-grained details of small targets, even in challenging scenarios with significant background clutter.
\begin{figure*}[!t]
    \centering
    \includegraphics[width=1.0\linewidth]{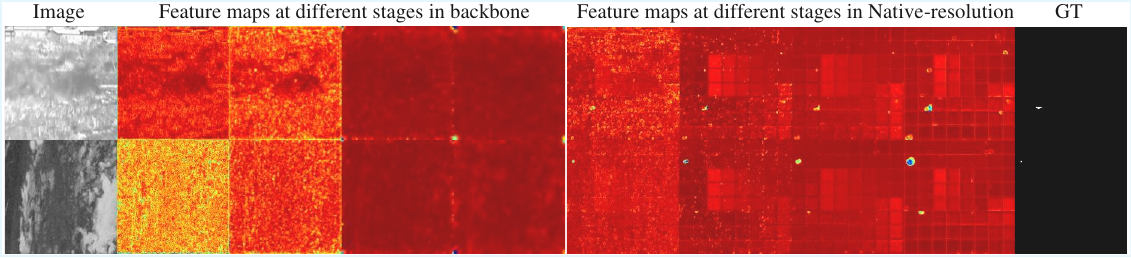}
        \caption{Visual comparison of feature maps in backbone encoder and native-resolution branch.}
    \label{fig:comp}
\end{figure*}

\subsection{Ablation Study}

\paragraph{Impact of the Native-Resolution Branch}
We visualize the features at different stages of the native-resolution branch and the backbone encoder, as shown in Figure \ref{fig:comp}. From the figure, it is evident that in images with small targets or complex backgrounds, the backbone encoder struggles to capture sufficient target features. As the image undergoes downsampling, the resolution of the target decreases, and the information related to the target gradually diminishes, making it harder for the model to maintain high-precision feature representations. This issue is particularly challenging when targets are small or obscured by complex background clutter.
In contrast, the native-resolution branch demonstrates a significant advantage. The features in this path progressively strengthen at each stage, retaining more fine-grained details throughout the network. This ability to preserve high-resolution information ensures that small or subtle target features are more easily captured, even in the presence of noise or background complexity.
By leveraging the native-resolution branch to extract native-resolution features, we supplement the backbone encoder with more precise target information, which not only improves the model's ability to detect small targets but also enhances its robustness in more challenging scenarios.

\rev{To more clearly evaluate the role of the native-resolution branch, we unify the ablation results into Table~\ref{table:nrb_comparison}, which jointly compares three settings: without the native-resolution branch, with a non-patch-based CNN branch, and with the proposed patch-based 2D+1D branch. As shown in the table, introducing the native-resolution branch consistently improves detection performance across all datasets, with the most notable gains observed on IRSTD-Hard, highlighting the importance of preserving fine-grained target details in challenging scenes. 
Among the compared branch designs, the patch-based 2D+1D architecture achieves the best overall performance across all datasets. By performing convolutional encoding within each patch and modeling inter-patch dependencies separately, it preserves fine-grained spatial details while maintaining global context awareness. The non-patch-based CNN variant performs slightly worse, which can be attributed to its limited capability in modeling long-range interactions under native-resolution processing. For completeness, we also attempted a no-downsampling ViT baseline, but it resulted in out-of-memory (OOM) errors even with batch size 1 on a single RTX 4090, indicating that dense ViT-based native-resolution modeling is prohibitively expensive in this setting. By contrast, the proposed native-resolution branch preserves the benefits of native-resolution processing while remaining computationally tractable.}

\begin{table*}[t]
\renewcommand\arraystretch{1.2}
\centering
\caption{\rev{Comparison of different native-resolution settings and architectures.}}
\begin{tabular}{l|ccc|ccc|ccc}
\toprule
\multirow{2}{*}{Dataset} 
& \multicolumn{3}{c|}{w/o native-resolution branch} 
& \multicolumn{3}{c|}{Non-Patch-based CNN} 
& \multicolumn{3}{c}{Patch-based 2D+1D (ours)} \\
& IoU & $P_d$ & $F_a$ 
& IoU & $P_d$ & $F_a$ 
& IoU & $P_d$ & $F_a$ \\
\midrule
IRSTD-1k   & 72.14 & 93.31 & 8.97  & 72.30 & 94.11 & 10.24 & 73.42 & 95.42 & 4.33 \\
SIRSTAUG   & 72.70 & 98.16 & 33.78 & 73.41 & 98.86 & 39.63 & 74.39 & 99.72 & 20.83 \\
NUDT       & 95.34 & 98.60 & 6.77  & 95.77 & 99.26 & 8.74  & 97.45 & 100.00 & 1.22 \\
IRSTD-Hard & 42.19 & 78.81 & 63.25 & 42.26 & 78.77 & 67.43 & 47.83 & 82.18 & 33.58 \\
\bottomrule
\end{tabular}

\label{table:nrb_comparison}
\end{table*}

\begin{figure}[t]
    \centering
    \includegraphics[width=1.0\linewidth]{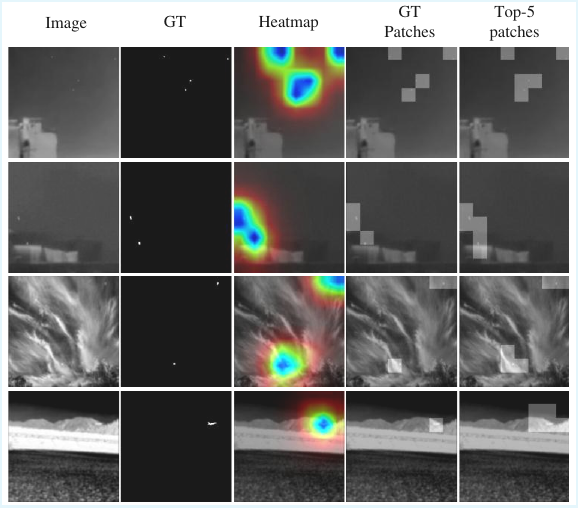}
        \caption{Visualization of patch-to-target relevance prediction.}
    \label{fig:patch to target}
\end{figure}

\paragraph{Impact of Two-stage Optimization and Soft Label}
\rev{As shown in Table~\ref{tab:joint_ablation}, the proposed two-stage strategy consistently outperforms one-stage end-to-end training under the same supervision type.  Under the extreme foreground sparsity of IRSTD, jointly learning patch relevance and dense segmentation from scratch is difficult, since the scoring MLP must distinguish a very small number of target-related patches from a large background-dominated patch set. As a result, one-stage optimization may produce unstable early patch rankings, weakening the effectiveness of native-resolution injection. 
In contrast, the proposed two-stage strategy first pretrains the native-resolution branch and scoring MLP to provide a more reliable initialization for patch ranking, and then fine-tunes the detector with the scoring MLP fixed, so that stable patch ranking can be used to guide native-resolution injection during segmentation training.
}

\begin{figure}[t!]
    \centering
    \includegraphics[width=1\linewidth]{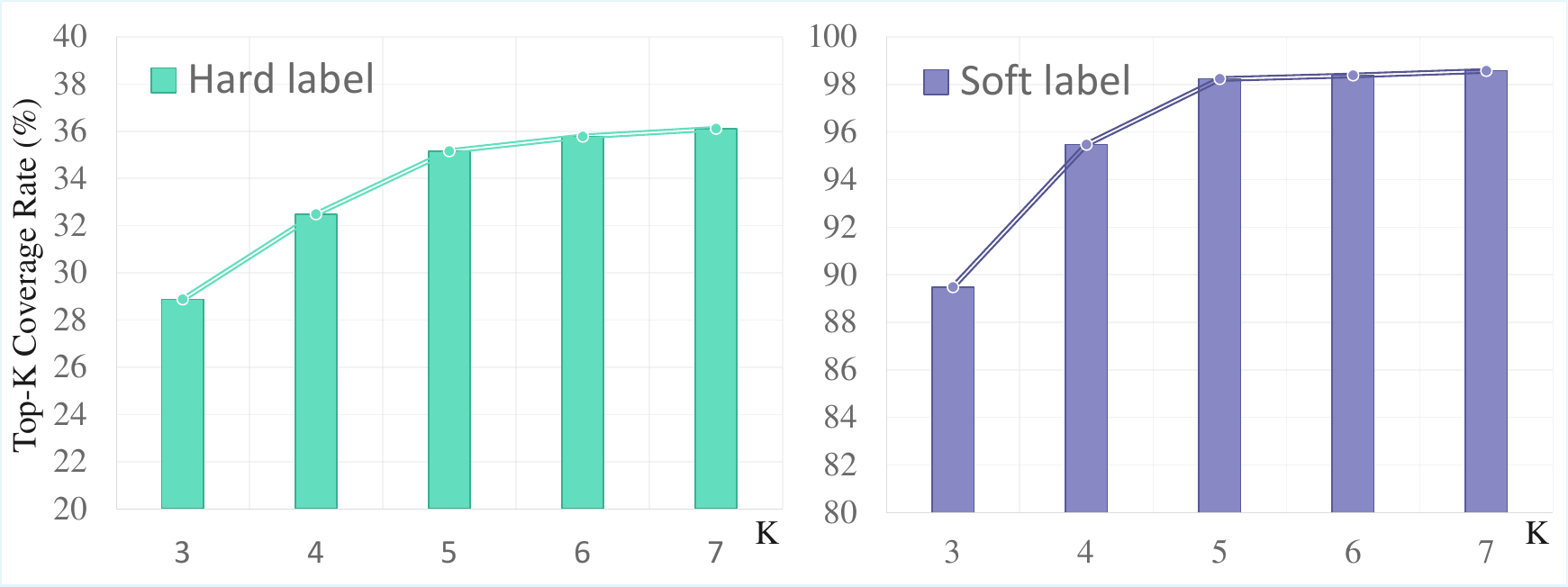}
        \caption{\rev{Top-$K$ patch coverage under two different {training} supervision strategies for the scoring network on NUDT-SIRST. Both curves are evaluated with the same coverage criterion at inference time.}}
    \label{fig:topk}
\end{figure}

\rev{The Top-$K$ coverage rate measures the proportion of selected patches that contain target regions. Fig.~\ref{fig:topk} compares two \emph{training} strategies for the scoring network, namely hard-label supervision and soft-label supervision, under the same coverage-based evaluation protocol. Soft-label training consistently achieves much higher coverage than hard-label training. This improvement can be attributed to the distance-aware supervision in Eq.~(\ref{eq:gaussian_label}), which alleviates extreme class imbalance and provides smoother gradient signals during relevance learning. To further evaluate the impact of supervision type and optimization strategy on the final detector, we compare four settings in Table~\ref{tab:joint_ablation}: one-stage training with hard-label supervision, one-stage training with soft-label supervision, two-stage training with hard-label supervision, and two-stage training with soft-label supervision. The results show that soft-label supervision consistently improves downstream detection under the same training strategy, while two-stage optimization further brings gains under the same supervision type. }

\paragraph{Impact of $K$}
\rev{We further analyze the influence of $K$ on detection performance. When $K=5$, the performance becomes nearly saturated, and increasing $K$ beyond this value yields only marginal gains. This is closely related to the target distribution of the evaluated datasets, where most images contain fewer than five targets. As shown in Fig.~\ref{fig:topk}, increasing $K$ from smaller values improves the coverage of target-relevant patches, while $K=5$ is already sufficient to preserve most useful native-resolution evidence on the current benchmarks. For scenes with substantially higher target density, a larger $K$ may be more appropriate, which deserves further investigation in future work. Under the default setting, the model retains only 5 out of 64 native patches while maintaining near-saturated detection performance.}

\begin{table*}[t]
\centering
\caption{\rev{Combined effect of training strategy and supervision type on downstream detection across four datasets.}}
\label{tab:joint_ablation}
\renewcommand\arraystretch{1.15}
\setlength{\tabcolsep}{4.5pt}
\begin{tabular}{l|ccc|ccc|ccc|ccc}
\toprule
\multirow{2}{*}{Dataset}
& \multicolumn{3}{c|}{One-stage + Hard-label}
& \multicolumn{3}{c|}{One-stage + Soft-label}
& \multicolumn{3}{c|}{Two-stage + Hard-label}
& \multicolumn{3}{c}{Two-stage + Soft-label } \\
& IoU & $P_d$ & $F_a$
& IoU & $P_d$ & $F_a$
& IoU & $P_d$ & $F_a$
& IoU & $P_d$ & $F_a$ \\
\midrule
IRSTD-1k   & 69.84 & 91.96 & 11.82 & 71.92 & 93.87 & 7.28  & 70.68 & 92.54 & 9.46  & 73.42 & 95.42 & 4.33 \\
SIRSTAUG   & 71.25 & 97.86 & 41.72 & 73.12 & 98.94 & 28.67 & 72.31 & 98.31 & 34.95 & 74.39 & 99.72 & 20.83 \\
NUDT       & 93.94 & 96.82 & 15.37 & 95.37 & 98.21 & 2.68  & 94.86 & 97.43 & 11.49 & 97.45 & 100.00 & 1.22 \\
IRSTD-Hard & 41.53 & 75.64 & 71.28 & 44.92 & 79.37 & 46.85 & 43.68 & 77.91 & 58.42 & 47.83 & 82.18 & 33.58 \\
\bottomrule
\end{tabular}
\end{table*}
\begin{figure}[t!]
    \centering
    \includegraphics[width=0.9 \linewidth]{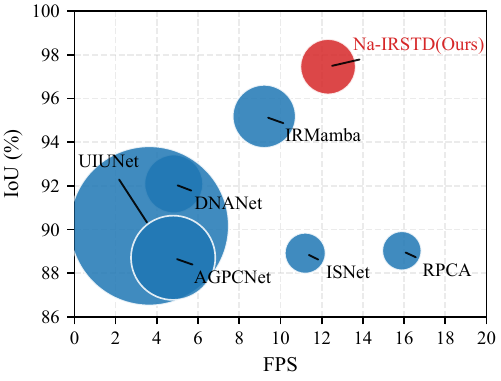}
        \caption{Comparison of IoU and throughput on a
single Nvidia GeForce 4090 GPU. The circle size refers to the
model size. Batch size is set to 1, and the experiments are
conducted on the NUDT dataset.}
    \label{fig:tradeoffs}
\end{figure}

\subsection{Trade-off Analysis}

To further assess the practical efficiency of Na-IRSTD,
we analyze the trade-off between detection accuracy and inference speed.
As shown in Figure~\ref{fig:tradeoffs}, Na-IRSTD achieves the highest detection accuracy
while maintaining competitive inference speed
and a moderate model size compared with existing methods.
In contrast, some methods achieve faster runtime
at the cost of reduced detection performance,
whereas others improve accuracy but require heavier models
or lower inference efficiency. These results indicate that the proposed native-resolution modeling
combined with selective token processing
effectively improves detection capability
without introducing disproportionate computational overhead.

\section{Conclusion}
In this article, we propose Na-IRSTD, a novel framework that fully leverages native-resolution representations for infrared small target detection. We introduce an effective native-resolution information extractor that preserves critical target details without the loss associated with downsampling. Additionally, the two-stage training strategy reduces training complexity, implements token reduction, and effectively filters redundant information while retaining key details. Experimental results on multiple datasets validate the effectiveness of Na-IRSTD under the evaluated settings.
\bibliographystyle{IEEEtran}
\bibliography{ref}

\end{document}